\documentclass{bmvc2k}

\usepackage{array}
\usepackage[table]{xcolor}
\usepackage{amssymb}
\usepackage{wrapfig}
\usepackage{booktabs}
\usepackage[flushleft]{threeparttable}

\title{Learning To Pay Attention To Mistakes}

\addauthor{Mou-Cheng Xu}{moucheng.xu.18@ucl.ac.uk}{1}
\addauthor{Neil P. Oxtoby}{n.oxtoby@ucl.ac.uk}{1}
\addauthor{Daniel C. Alexander}{d.alexander@ucl.ac.uk}{1}
\addauthor{Joseph Jacob}{j.jacob@ucl.ac.uk}{1}
\addinstitution{
 Centre For Medical Image Computing\\
 Department of Computer Science\\
 Department of Medical Physics \& Biomedical Engineering\\
 University College London \\
 London, UK
}

\runninghead{Xu, Oxtoby, Alexander \& Jacob}{Learning To Pay Attention To Mistakes}

\begin{document}

\maketitle

\begin{abstract}
In convolutional neural network based medical image segmentation, the periphery of foreground regions representing malignant tissues may be disproportionately assigned as belonging to the background class of healthy tissues \cite{attenUnet}\cite{AttenUnet2018}\cite{InterSeg}\cite{UnetFrontNeuro}\cite{LearnActiveContour}. Misclassification of foreground pixels as the background class can lead to high false negative detection rates. In this paper, we propose a novel attention mechanism to directly address such high false negative rates, called Paying Attention to Mistakes. Our attention mechanism steers the models towards false positive identification, which counters the existing bias towards false negatives. The proposed mechanism has two complementary implementations: (a) ``explicit'' steering of the model to attend to a larger Effective Receptive Field on the foreground areas; (b) ``implicit'' steering towards false positives, by attending to a smaller Effective Receptive Field on the background areas. We validated our methods on three tasks: 1) binary dense prediction between vehicles and the background using CityScapes; 2) Enhanced Tumour Core segmentation with multi-modal MRI scans in BRATS2018; 3) segmenting stroke lesions using ultrasound images in ISLES2018. We compared our methods with state-of-the-art attention mechanisms in medical imaging, including self-attention, spatial-attention and spatial-channel mixed attention. Across all of the three different tasks, our models consistently outperform the baseline models in Intersection over Union (IoU) and/or Hausdorff Distance (HD). For instance, in the second task, the ``explicit'' implementation of our mechanism reduces the HD of the best baseline by more than $26\%$, whilst improving the IoU by more than $3\%$. We believe our proposed attention mechanism can benefit a wide range of medical and computer vision tasks, which suffer from over-detection of background.
\end{abstract}
%------------------------------------------------------------------------- 
\section{Introduction}
\label{sec:intro}
Convolutional Neural Networks (CNN) enhanced by attention mechanisms have recently been transferred from computer vision to medical image analysis to tackle segmentation tasks \cite{AttenUnet2018, CSEUnet2018}. Attention mechanisms aim to focus learning on salient regions of interest (RoI), i.e. foreground pixels in medical images, to minimise the misclassification of RoI. Attention is implemented through a normalisation step in the latent feature space to focus the network on the RoI. However, foreground pixels in medical images are often heavily under-represented. The resulting bias towards detection of background areas causes an under-detection of foreground pixels (True Positives: TPs) and an over-detection of background pixels (False Negatives: FNs), especially at the edges of a RoI (see visual results in \cite{attenUnet, AttenUnet2018, InterSeg, UnetFrontNeuro, LearnActiveContour}). It is apparent that reducing FN detection is a key challenge to increasing the utility of CNN based segmentation.

\begin{wrapfigure}{r}{0.4\textwidth}
  \vspace{-30 pt}
  \begin{center}
    \includegraphics[width=40mm]{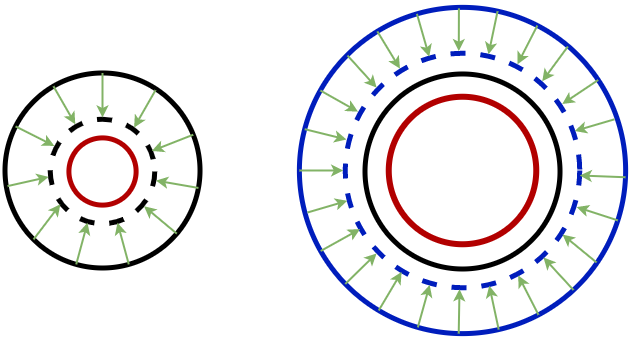}
  \end{center}
  \vspace{-10 pt}
  \caption{Left: Performance illustration of the existing attention mechanisms. Right: Performance illustration of our proposed solution to reduce FNs.}
  \vspace{-10 pt}
\label{fig:Performance_upper_bound}
\end{wrapfigure}
% \textcolor{red}{todo: change figure 1 with a real example, also change this following paragraph}
Attention is a natural mechanism by which TP detection can be improved. Yet existing attention mechanisms do not adequately reduce FNs (see section \ref{sec:results}). Existing attention mechanisms focus on the labelled foreground areas as the RoI (area within the black circle in \textbf{Figure} \ref{fig:Performance_upper_bound}), which we refer to as \textit{Paying Attention to True Positives (TPs)}. Ideally, a well-trained over-parameterized model is able to detect the entirety of the RoI, but the bias towards detecting background pixels leads to the area of focus shrinking (dotted black circle in \textbf{Figure} \ref{fig:Performance_upper_bound}, Left). This is exacerbated by the optimisation towards local minimum solutions which results in further shrinkage of the regions identified as TPs (red circle in \textbf{Figure} \ref{fig:Performance_upper_bound}, Left).

To overcome the high FN rate inherent to existing attention mechanisms, we propose an alternative strategy, which we refer to as \textit{Paying Attention To Mistakes}. The hypothesis is that FNs can be reduced by encouraging a bias towards False Positives (FPs) detection, focused particularly around the boundary of TPs. This is shown conceptually in \textbf{Figure} \ref{fig:Performance_upper_bound} (Right), where our proposed attention mechanism learns to focus on an expanded RoI (blue solid circle). After shrinkage occurs, as explained in the last paragraph, resulting in more TPs, the red circle in Figure 1, Right is now more closely approximated to the black circle, representing the ground truth RoI. The black circles in Figure 1 (left,right) are identical in size. \textit{Paying Attention To Mistakes} requires neither extra annotations of FPs, nor modifications to the source data distribution, which might risk information loss. Our main \textbf{contributions} are three-fold:
\begin{description}\setlength\itemsep{-2mm}
  \item[$\bullet$] We are the first to use an attention mechanism to ameliorate the pixel-wise classification bias towards false negatives in medical imaging.  
  \item[$\bullet$] We are the first to develop an attention mechanism based on the Effective Receptive Field (ERF). This was implemented to make our attention mechanism ``transparent''. Our mechanism has two complimentary implementations to \textit{Pay Attention To Mistakes} which are explicit and implicit respectively.
  \item[$\bullet$] We perform extensive experiments including comparisons with state-of-the-art baselines and ablation studies on different configurations on three different data sets.
  \label{contribution}
\end{description}

\section{Related works}
\label{sec:related_works}
\textbf{Attention mechanisms.} We review a few representative attention mechanisms that \textit{Pay Attention To TPs} here. According to the focusing target (e.g.~what to attend to or where to attend), most existing attention mechanisms can be divided into three groups: channel attention (e.g.~importance of each channel of a feature map) \cite{SE2018}, spatial attention (e.g.~importance of each spatial location of a feature map) \citep{LearnAtten2018, Bilinear2015, NonLocal2018} and spatial-channel mixed attention (e.g. importance of each spatial location at each channel of a feature map) \citep{ResAtten2017, CBAM2018}. Channel attention is based on weighting each channel according to each channel's most representative feature (e.g.~mean \cite{SE2018}, maximum or both \cite{CBAM2018}). In spatial attention, self-attention mechanisms have been used as in \citep{Bilinear2015, NonLocal2018}; or semantic features from deep layers have been used as ``keys'' to enhance representation learning in shallow layers as in \citep{LearnAtten2018, ResAtten2017}.

\textbf{Effective receptive field.} In a layer of a CNN, the size of the region corresponding to the neuron in the next layer is called the ``receptive field'' (RF). The centre of the RF has the highest impact on the layer output. The impact of the pixels across the RF has been shown to resemble a Gaussian distribution. Therefore, in a forward pass, the gradient of the signal decays from the RF centre to the periphery in a squared exponential manner \cite{ERF}. Accordingly, only a fraction of the RF is detected and contributes to the output. This effective area is called the ``Effective RF'' (ERF). The ERF also reduces \cite{ERF} as the network goes deeper. More importantly, it has been shown that the size of an ERF is also influenced by neural network topology \cite{ERF}. By combining the ERF with latent space embedding \cite{CAM2016}, we can achieve flexible control of the area over which we want the model to focus.

\textbf{Other related work.} Dilated convolutional layers \cite{Deeplab} have been proposed to expand the receptive field in supervised image segmentation tasks. Work also has previously been proposed for reversed attention (RA) \citep{ReverseAttentionECCV, ReverseAttentionBMVC} to improve classification accuracy in confusing regions. Our methods differ from related works in both motivation and implementation. We are motivated to use an attention mechanism to expand or shrink the focus on the RoI, to control detection biases at decision boundaries, whereas previous works are motivated to eliminate the bias. Regarding implementation, our approach is the first attention mechanism built upon the ERF. 

\section{Methods}
\label{sec:methods}
Our research hypothesis is that the bias towards FNs detection in medical image segmentation could be mitigated by forcing networks to favour FPs detection using an attention mechanism. It is possible to ``explicitly'' shift the bias at decision boundaries towards detection of FPs, termed ``False Positive Attention''. It is also possible to reverse labels for foreground and background pixels and ``implicitly'' focus on detecting the foreground. The two candidates can result in complimentary implementations. We first present the implementation which ``explicitly'' shifts the bias towards detection of FPs.
\subsection{False Positive Attention (FPA)}
To ``explicitly'' shift the bias towards FPs, we use the strategy as explained in \textbf{Figure} \ref{fig:Performance_upper_bound}. We expand the focus of the model beyond the area of labelled TPs, to extend into regions of FPs by applying a convolutional layer \cite{Deeplab} to learn a larger smoothed ERF. The larger ERF extends from the original ERF, guided by a dilated convolutional layer at the same depth. The rationale behind the architecture of FPA is based on one empirical result in \cite{ERF}: at the same depth, the ERF of the dilated convolutional layer is larger than the ERF of the original convolutional layer. To merge the information contained in the regions surrounding the ERF to the ERF, it would be intuitive to use the mean of the the outputs of the convolutional and dilated convolutional layers. However this mean calculation eliminates the bias towards FP detection \cite{DilationSemiSupervised}. We therefore use a Sigmoid function on the output of the dilated convolutional layer to create a smoothed larger ERF, before performing an element-wise multiplication on the output of the corresponding convolutional layer. 

\begin{wrapfigure}{l}{0.4\textwidth}
  \vspace{-10 pt}
  \begin{center}
    \includegraphics[width=0.35\textwidth]{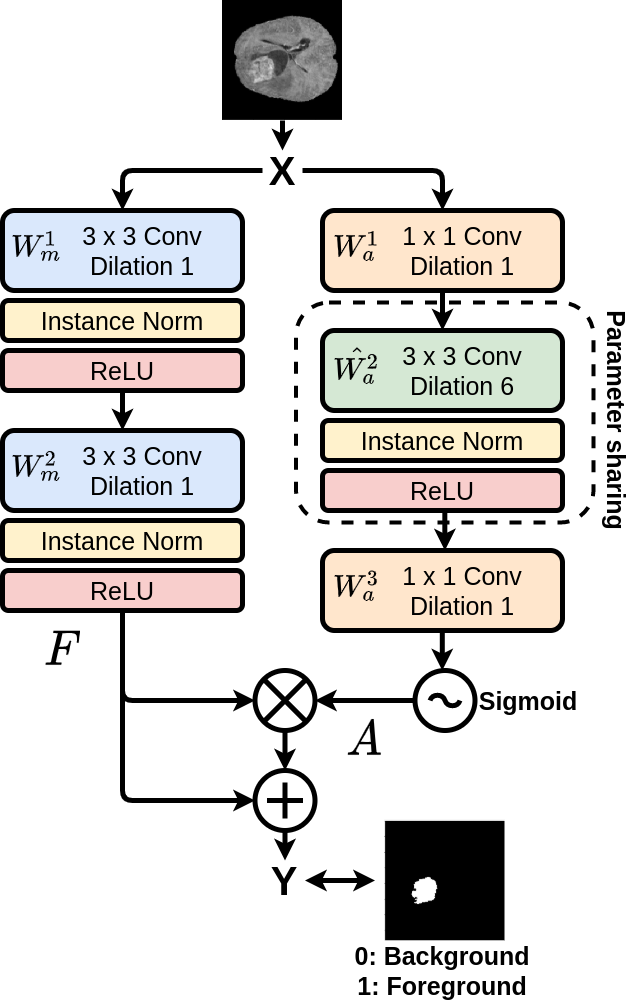}
  \end{center}
  \vspace{-20 pt}
  \caption{False Positive Attention}
  \vspace{-40 pt}
\label{fig:fpa}
\end{wrapfigure}
Hence, our approach whereby we shift the bias of the model towards FP detection represents an attention mechanism. We now describe the operation of FPA. The FPA module consists of two parallel branches: the main branch processes visual information, whereas the attention branch generates the smoothed enlarged ERF. Given the input feature map $X \in R^{C \times H \times W}$, the output feature map of main branch is $F \in R^{2C \times H/2 \times W/2}$ and the output attention weights of the attention branch is $A \in R^{2C \times H/2 \times W/2}$. We achieve the output feature map ($Y \in R^{2C \times H/2 \times W/2}$) following the equations:
\begin{equation}
\begin{aligned}
    F = W_{m}^{2}[W_{m}^{1}(X)] \\
    A = \sigma(W_{a}^{3}\{\hat{W_{a}^{2}}[W_{a}^{1}(X)]\}) \\
    Y = F \odot A + F
\end{aligned}
\label{equation_fpa}
\end{equation}
Where $W_m^{i}$ denotes the $i^{th}$ standard convolutional layer in the main branch; $\hat{W_a^{i}}$ denotes the $i^{th}$ dilated convolutional layer in the attention branch; $W_a^{i}$ denotes the $i^{th}$ standard convolutional layer in the attention branch; $\sigma$ is Sigmoid function; $\odot$ is element-wise multiplication. All of the Non-linear activation and normalisation layers are omitted for simplicity of expression. The detailed architectures can be found in \textbf{Figure} \ref{fig:fpa}.
% Directly reducing False Negative identification by biasing the model decision boundary ``away'' from False Negatives is challenging. 
%  
\subsection{Reverse False Negative Attention (RFNA)}
On the contrary to FPA which directly favours FP detection, we present an alternative approach to reduce over-detection of background by favouring ``reverse FN detection''. It is obvious that if we directly bias towards FN, we risk in deteriorating the performance which already suffers from high FN rate. Our solution is to ``implicitly'' \textit{Pay Attention to Mistakes}. In binary segmentation, the labels values normally are: 1 for foreground and 0 for background, and the models would naturally favour FNs detection (section \ref{sec:intro}) where foreground pixels are classified as label value 0. Our implementation first reverses the labels values whereby the models would naturally favour reverse FP detection, which is classification of foreground with a label value of 1. We now encourage the models to focus less on reverse FP detection, which is equivalent to biasing towards reverse FNs detection. This bias towards reverse FNs in RFNA share the same goal with the bias towards FPs in FPA, which is to reduce false detection of foreground areas as background class, albeit RFNA reduces over-detection of background in an ``implicit'' way.

As this implementation favours reverse FNs detection, it is termed ``Reverse False Negative Attention (RFNA)''.The RFNA module has two branches but these are different from the parallel architecture in FPA. The attention branch in RFNA is placed sequentially after the main branch. The rationale behind the architecture of RFNA is based on another empirical observation in \cite{ERF}: the RF increases linearly with network depth, while the ratio between the ERF and the RF rapidly decreases. This phenomenon leads to a situation where a proximal deeper layer might have a smaller ERF than a shallower layer. For instance, the 40th convolutional layer would have a smaller ERF than the 20th layer as shown in \textbf{Figure 1} in \cite{ERF}. Additionally, it was found in \cite{ERF} that residual connections also generate a smaller ERF. 

\begin{wrapfigure}{l}{0.3\textwidth}
  \vspace{-15 pt}
  \begin{center}
    \includegraphics[width=0.2\textwidth]{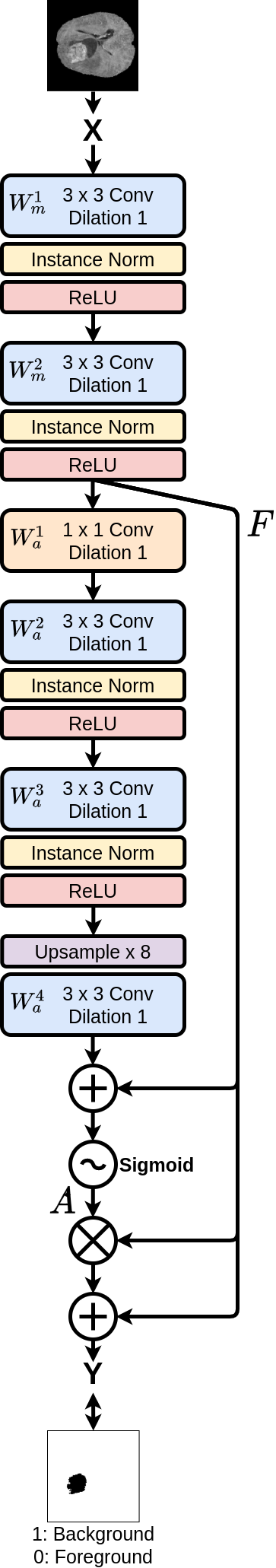}
  \end{center}
  \vspace{-10 pt}
  \caption{Reverse False Negative Attention}
  \vspace{-100 pt}
\label{fig:fna}
\end{wrapfigure}
We combine these two characteristics to generate a smaller ERF to guide the networks to shrink the focus on reverse FP. We use the same notations from FPA to describe the operation of RFNA as follows:
\begin{equation}
\begin{aligned}
    F = W_{m}^{2}[W_{m}^{1}(X)] \\
    A = \sigma \{W_{a}^{4}[UpSample(W_{a}^{3}\{W_{a}^{2}[W_{a}^{1}(F)]\})] + F\} \\
    Y = F \odot A + F
\end{aligned}
\label{equation_rfna}
\end{equation}
Where $UpSample$ denotes a bilinear upsampling layer, the default upsampling ratio is 8 (\textbf{Figure} \ref{fig:fna}). We use the output of the main branch as the input for the attention branch to make sure that we use deeper features, leading to a smaller ERF. There is an additional residual connection before the Sigmoid function, which aids generation of a smaller ERF. The architecture of RFNA is in \textbf{Figure} \ref{fig:fna}.

\subsection{Implementation}
\label{sec:implementation}
We enhance a 3 down-sampling stage, 2D U-net \cite{Unet} as our backbone, by replacing the batch normalisation with instance normalisation \cite{InstanceNorm} and replacing deconvolutional layers with bilinear upsampling layers. We replace convolutional blocks in the encoder in U-net with the proposed attention modules. The channel number in the first encoder in the backbone is 32, as we found the original number 64 is redundant for our tasks. In FPA, $\hat{W_{a}^2}$ is repeated twice via parameter sharing. In RFNA, $W_{a}^1$ expands the channel number to 4 times of the channel number of output; $W_{a}^2$ and $W_{a}^3$ use depth-wise convolution layers for computational efficiency. Ablation studies on different configurations of these components are outlined in a later section \ref{sec:ablation}.

\section{Experiments}
\subsection{Materials} 
\textbf{CityScape} Our first task is a synthetic medical task using high-resolution RGB images from CityScapes \cite{cityscape}. We perform a binary segmentation task to distinguish between vehicles and the background, replicating medical image segmentation tasks which are typically dense binary predictions. Due to computational restrictions, we downsample images to 256x128. We use cities from ``train'' and ``val'' for training and testing, respectively. We hold one city for validation in training data.

\textbf{BRATS} The publicly available BRATS 2018 Training data \cite{brats} has 210 High Grade Glioma (HGG) cases and 76 Low Grade Glioma (LGG) cases. Since the enhancing tumour core is the hardest class to discern \cite{BRATS2018Winner, BRATS2018Second}, we focus on this class to illustrate the effectiveness of our methods. We noticed that LGG cases contain almost no enhancing tumour cores \cite{BRATS2018Second}. Therefore to focus on assessing the network, we only analysed the HGG cases. Pre-processing steps included: normalisation of each case for each modality; centre cropping to 160x160; concatenation of each modality as 4D input. 5 fold cross-validation with a case-wise split.

\textbf{ISLES} The ISLES2018 \cite{isles2018} training data contains 94 acute stroke CT perfusion scans. We randomly split the cases at the ratio of 0.7:0.1:0.2 for training, validation and testing. We use CBF, MTT, CBV, TMAX modalities, and normalise each case in each modality and centre crop them to 224x224. Our task was to segment the stroke lesions in the CT brain images.

\subsection{Baselines}
We compare our models with existing attention mechanisms in medical imaging, which \textit{Pay Attention To TPs}. We follow the categories in section \ref{sec:related_works} to select our baselines. For spatial attention, we use the state-of-the-art ``Attention U-net'' \cite{attenUnet}, denoted as ``AUnet''. For self-attention, we use a state-of-the-art efficient non-local module ``GCN'' \cite{GCNet}, to avoid the high computational burden of the original self-attention module \cite{NonLocal2018} and we integrate them into the backbone as ``GCNUnet''. The backbone is described in section \ref{sec:implementation}.
\begin{wraptable}{l}{0.52\textwidth}
  \vspace{-10 pt}
  \caption{Where the convolutional blocks are replaced with attention modules in baselines and our networks. Type: attention category.}\label{Table_main}%
  \vspace{-15 pt}
  \begin{center}
    \begin{tabular}{l l l l}
    \hline
    \bfseries Networks & \bfseries Type & \bfseries Encoder & \bfseries Decoder \\
    \hline
    AUnet & spatial &  &\checkmark\\
    CSCUnet & mixed &\checkmark & \checkmark\\
    CBAMUnet & mixed &\checkmark & \\
    GCNUnet & self &\checkmark & \\
    \textbf{FPA/RFNA} & mixed &\checkmark & \\
    \hline
    \end{tabular}
  \end{center}
  \vspace{-20 pt}
\label{table:position_attention}
\end{wraptable}
For spatial-channel mixed attention, we first use the state-of-the-art mixed attention U-net called ``concurrent spatial and channel squeeze-excitation U-net'' \cite{CSEUnet2018}, this is denoted as ``CSCUnet''. We also include another state-of-the-art mixed attention mechanism, namely ``CBAM'' \cite{CBAM2018} and we also implement them into the backbone as ``CBAMUnet''. No channel attention is included, as ``CSC'', ``CBAM'', ``GCN'' already comprise the state-of-the-art channel attention \cite{SE2018}. In implementation of baselines, the convolutional blocks in the backbone are replaced with different existing attention modules, either following the references or in the encoder as we insert our FPA/RFNA in the encoder, see details in \textbf{Table} \ref{table:position_attention}.

\subsection{Ablation Studies}
\label{sec:ablation}
\textbf{Effect of attention branches} We study the effect of our attention modules by pruning the branches in FPA/RFNA. As shown in \textbf{Figure} \ref{fig:fpa}, without the main branch, we have a U-net with dilated convolutional layers in the encoder. We replace the convolutional layers in the encoder in the backbone with dilated convolutional layers as ``D6/9 Unet'' (6 or 9 is the dilation rate). Without the attention branch in \textbf{Figure} \ref{fig:fpa}, it becomes the backbone Unet. Similarly, we also have Unet without the attention branch in RFNA.

\textbf{Effect of downsampling ratio in RFNA} We prune the convolutoinal layers in RFNA to study how the downsampling ratio in the attention branch influences performance. We first remove $W_{a}^3$ and use $UpSample \times 4$ to see the effect of a downsampling ratio of 4. Then we further remove $W_{a}^2$ and use $UpSample \times 2$ to check the effect of a downsampling ratio of 2. 

\textbf{Effect of dilation rate in FPA} The dilation rate in FPA is an important hyper-parameter, so we compare different dilation ratios at 6, 9 and 12. No smaller dilation is used as we found that a dilation ratio 3 has no obvious effect on segmentation results. 

\textbf{Effect of model capacity} To examine whether the effect of our attention modules was a consequence of accumulating more parameters, we double the channel number in the backbone to make a wide U-net, denoted as ``WUnet''. 

\textbf{Effect of channel numbers} We study the impact of model capacity in both FPA and RFNA. For each configuration in FPA with a different dilation rate and each configuration in RFNA with a different downsampling ratio, two variants using depth-wise convolutional layers and no depth-wise convolutional layers in attention branches are implemented. For the variant of using depth-wise convolutional layers, we explore the impact of channel expansion ratio at 2, 4 and 8 of output channel number, in $W_{a}^1$. For the variant of using no depth-wise convolutional layers, we explore the impact of channel expansion ratio at 1 and 2, due to computational restriction. In default (section \ref{sec:implementation}), FPA uses no depth-wise convolutional layers and RFNA uses depth-wise convolutional layers.
\subsection{Training}
\begin{wraptable}{r}{0.5\textwidth}
  \vspace{-12 pt}
  \caption{Training details. lr: learning rate.}\label{Table_main}%
  \vspace{-5 pt}
  \begin{center}
    \begin{tabular}{l c c c}
    \hline
    \bfseries Dataset & \bfseries Epoch & \bfseries Batch & \bfseries lr\\
    \hline
    CityScapes & 100 & 8 & 2e-4\\
    BRATS & 80 & 50 & 1e-4\\
    ISLES & 60 & 80 & 1e-3\\
    \hline
    \end{tabular}
  \end{center}
  \vspace{-20 pt}
\label{table:training}
\end{wraptable}
AdamW optimiser \cite{adamw} is used for optimisation. Dice Loss \cite{vnet} is used as objective function. Random horizontal flipping is used for augmentation. Training details are in \textbf{Table} \ref{table:training}. No further improvements were observed with more training epochs in our settings. All experiments were run for at least 3 times on a NVIDIA TITAN V GPU. Our demo code is implemented in Pytorch 1.0 and it is available in:\url{https://github.com/moucheng2017/Pay_Attention_To_Mistakes}.

\section{Results}
\label{sec:results}

\begin{table}[!ht]
\vspace{-5 pt}
\centering
\begin{minipage}[t]{0.4\linewidth}\centering
\caption{Results on CityScapes.}
\label{table:result_cityscapes}
\begin{tabular}{l | c c}
  \hline
  \bfseries Networks & \bfseries IoU ($\%$) & \bfseries HD \\
  \hline
  AUnet & 54.17 $\pm$ 0.15 & 64.69 $\pm$ 0.67\\
  CSCUnet & 54.45 $\pm$ 0.31 & 60.15 $\pm$ 1.96\\
  \hline 
    % \bfseries Ours & \bfseries IoU (\%) & \bfseries HD \\
%   \hline
   \textbf{RFNA} & 55.62 $\pm$  0.45 & 65.18 $\pm$ 3.36\\
   \textbf{FPA} & \textcolor{red}{\textbf{59.39}} $\pm$ \textcolor{red}{\textbf{0.65}} & \textcolor{red}{\textbf{49.10}} $\pm$ \textcolor{red}{\textbf{2.51}}\\
  \hline
\end{tabular}
\end{minipage}\hfill%
\begin{minipage}[t]{0.5\linewidth}\centering
\caption{Results on ISLES2018.}
\label{table:result_isles}
\begin{tabular}{l | c c}
  \hline
  \bfseries Networks & \bfseries IoU ($\%$) & \bfseries HD \\
  \hline
  Unet & 52.41 $\pm$ 0.78 & 25.49 $\pm$ 1.38\\
  AUnet & 52.35 $\pm$ 0.63 & 23.66 $\pm$ 0.87\\
  CSCUnet & 52.70 $\pm$ 0.69 & 27.28 $\pm$ 0.72\\
  \hline 
    % \bfseries Ours & \bfseries IoU (\%) & \bfseries HD \\
%   \hline
  \textbf{RFNA} & 52.99 $\pm$ 1.63 & \textcolor{red}{\textbf{18.10}} $\pm$ \textcolor{red}{\textbf{4.75}}\\
  \textbf{FPA} & \textcolor{red}{\textbf{54.61}} $\pm$ \textcolor{red}{\textbf{0.54}} & 19.44 $\pm$ 1.41 \\
  \hline
  \end{tabular}
\end{minipage}
\vspace{-10 pt}
\end{table}

\begin{table*}[!ht]
\vspace{0 pt}
\centering
\caption{Results on BRATS2018. HD: Hausdorff distance at 95$\%$ percentile. FP: False Positive Rate. FN: False Negative Rate. Param: Parameters.}
\centering
\begin{tabular}{l | c c c c c}
  \hline
  \bfseries Networks & \bfseries IoU ($\%$) & \bfseries HD & \bfseries FP (\%) & \bfseries FN ($\%$) & \bfseries Param (M)\\
  \hline
  AUnet & 66.48 $\pm$ 1.04 & 15.51 $\pm$ 0.82 & $<$1e-2 & 58.42 $\pm$ 2.58 & 3.16\\
  CSCUnet & 66.75 $\pm$ 1.00 & 15.50 $\pm$ 1.79 & $<$1e-2 & 58.11 $\pm$ 3.28 & 3.14\\
  GCNUnet & 66.57 $\pm$ 1.20 & 15.11 $\pm$ 1.05 & $<$1e-2 & 58.16 $\pm$ 2.19 & 3.15\\
  CBAMUnet & 66.52 $\pm$ 1.50 & 15.19 $\pm$ 1.59 & $<$1e-2 & 58.68 $\pm$ 3.11 & 3.14\\
  \hline 
    % \bfseries Ours & \bfseries IoU (\%) & \bfseries HD & \bfseries FP (\%) & \bfseries FN (\%) & \bfseries Param (M)\\
%   \hline
  \textbf{RFNA} & \textcolor{red}{\textbf{71.15}} $\pm$ \textcolor{red}{\textbf{1.78}} & 21.76 $\pm$ 1.08 & $<$1e-2 & 50.15 $\pm$ 2.94 & \textcolor{red}{\textbf{5.63}}\\
  \textbf{FPA} & 70.10 $\pm$ 2.06 & \textcolor{red}{\textbf{10.86}} $\pm$ \textcolor{red}{\textbf{1.66}} & $<$1e-2 & \textcolor{red}{\textbf{49.39}} $\pm$ \textcolor{red}{\textbf{4.56}} & 4.2\\
  \hline
  \end{tabular}
  \vspace{-10 pt}
  \label{table:result_brats}
\end{table*}
Our \textit{Paying Attention To Mistakes} outperforms all of the baselines across three different multi-modal data sets in either Intersection over Union (IoU) or Hausdorff Distance (HD) or both. In the first task, RFNA improves the best baseline by 1.17\% in IoU; whereas FPA improves the IoU by 4.94\% and reduces the Hausdorff Distance (HD) of the best baseline by 18\%. In the second task, the FPA reduces the HD of the best baseline by 28.12\%. While the RFNA achieves the highest IoU score, with a 4.4\% margin compared to the best baseline. In the third task, despite the fact that perfusion CT images are challenging to interpret, both FPA and RFNA outperform the baselines in both accuracy metrics. RFNA reduces the HD by 23.49\% of the best baseline and FPA has a 2.26\% gain in IoU. 

\begin{wrapfigure}{r}{0.4\textwidth}
  \vspace{-25pt}
  \begin{center}
    \includegraphics[width=0.4\textwidth]{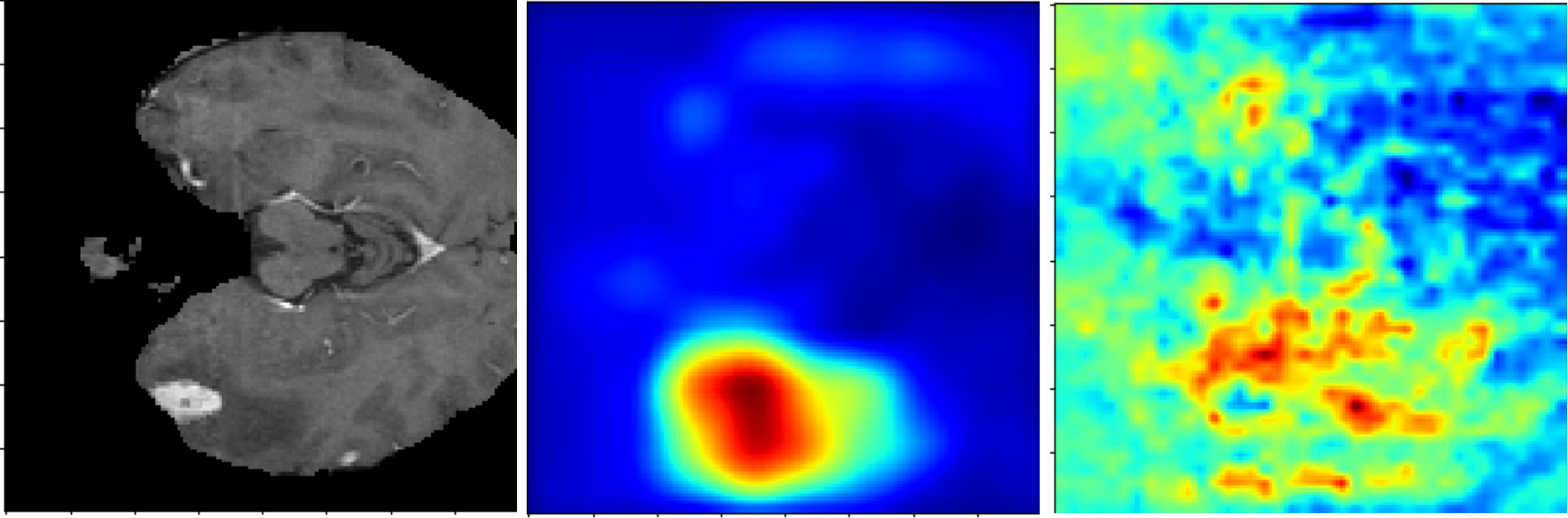}
  \end{center}
  \vspace{-20 pt}
  \caption{Left: Input (BRATS). Middle: RFNA. Right: FPA.}
  \vspace{-10 pt}
\label{fig:atten_visulisation}
\end{wrapfigure}
FPA consistently improved performance in both accuracy metrics across all three tasks. Although RFNA also consistently improved performance as measured by IoU, it has mixed impacts in HD. HD and IoU are good at measuring different types of mistakes, suggesting that FPA and RFNA actually refine the segmentation in different ways. To evaluate this further we visualised the attention maps in FPA and RFNA of the foreground class in the deepest encoder in \textbf{Figure} \ref{fig:atten_visulisation}. As demonstrated in \textbf{Figure} \ref{fig:atten_visulisation}, FPA and RFNA utilise global and local spatial information, respectively.
\begin{wrapfigure}{l}{0.5\textwidth}
  \vspace{-22pt}
  \begin{center}
    \includegraphics[width=0.5\textwidth]{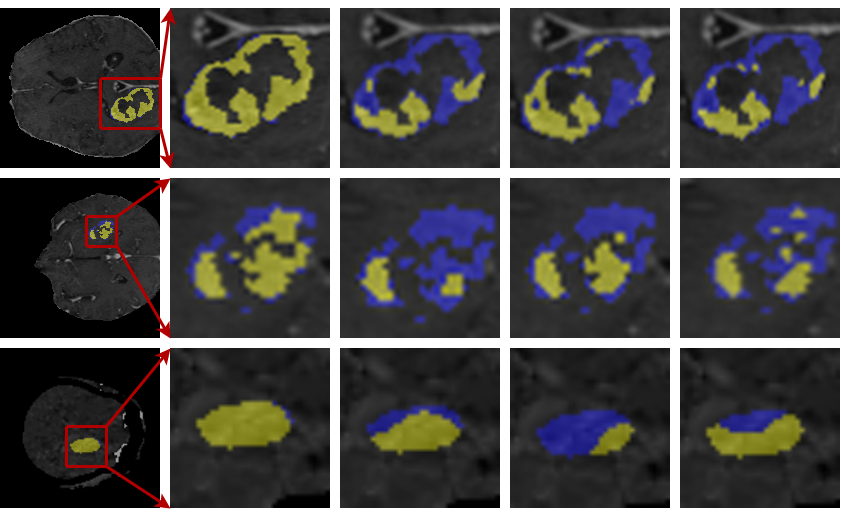}
  \end{center}
  \vspace{-20 pt}
  \caption{Blue: False Negatives, Yellow: True Positives. Row 1\&2: BRATS; row3: ISLES. 1st column: FPA, 2nd column: FPA (Zoomed in). 3rd column: AUNet, 4th column: CSCUnet, 5th column: UNet.}
  \vspace{-40 pt}
\label{fig:visual_brats}
\end{wrapfigure}
FPA and RFNA also experience different levels of bias in segmentation outcomes due to label reversal. Also, the ratios between background and foreground are different in each data set. Eventually, the bias differences in the data plus the mechanistic differences led to different performances between FPA and RFNA.

\begin{figure}[!ht]
  \vspace{0 pt}
  \centering
  \includegraphics[width=\textwidth]{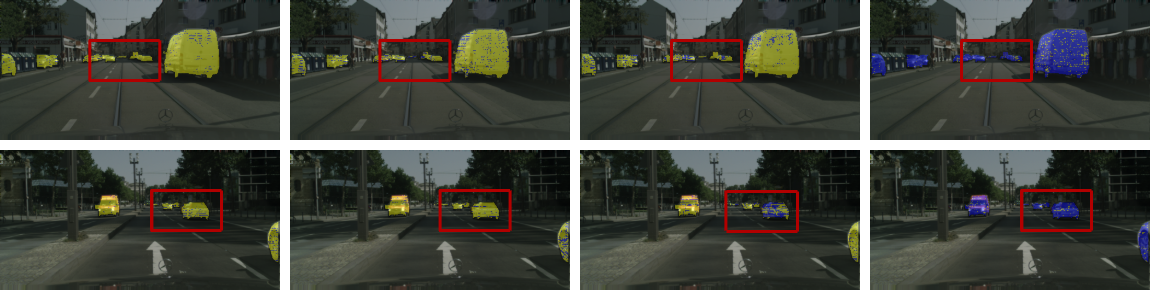}
  \caption{Blue: False Negatives, Yellow: True Positives. Visual results on Cityscapes. 1st column: FPA, 2nd column: RFNA, 3rd column: AUnet, 4th column: CSCUNet.}
  \vspace{0 pt}
\label{fig:visual_cityscape}
\end{figure}

The inferior performance of the baselines might be a result of their focus on TP regions, as discussed in section \ref{sec:intro}. By \textit{Paying Attention To Mistakes}, we successively reduce FN detection as qualitatively shown in \textbf{FN} column in \textbf{Table} \ref{table:result_brats}, where FPA reduces the FN rate of the best baseline by 15\%. The reduction of FN detection can also be seen in \textbf{Figure} \ref{fig:visual_brats} and \textbf{Figure} \ref{fig:visual_cityscape}.

% \textcolor{red}{todo: add explanation of why hd performance differs}

\subsection{Results on ablation studies}
Both FPA/RFNA in \textbf{Table} \ref{table:result_brats} performed better than backbone ``Unet'' in \textbf{Table} \ref{table:result_ablation}, suggesting that attention branches improve the performance of the backbone. We find that the combination of an attention branch and a main branch in FPA outperforms the use of an attention branch alone, as both ``D6Unet'' and ``D9Unet'' in \textbf{Table} \ref{table:result_ablation} performed worse than FPA. We also show that the positive effect of our attention mechanisms is not just from a larger number of parameters, as ``WUnet'' (\textbf{Table} \ref{table:result_ablation}) achieves a lower performance despite having three times more parameters than FPA.

\begin{figure}[!ht]
  \vspace{0 pt}
  \centering
  \includegraphics[width=\textwidth]{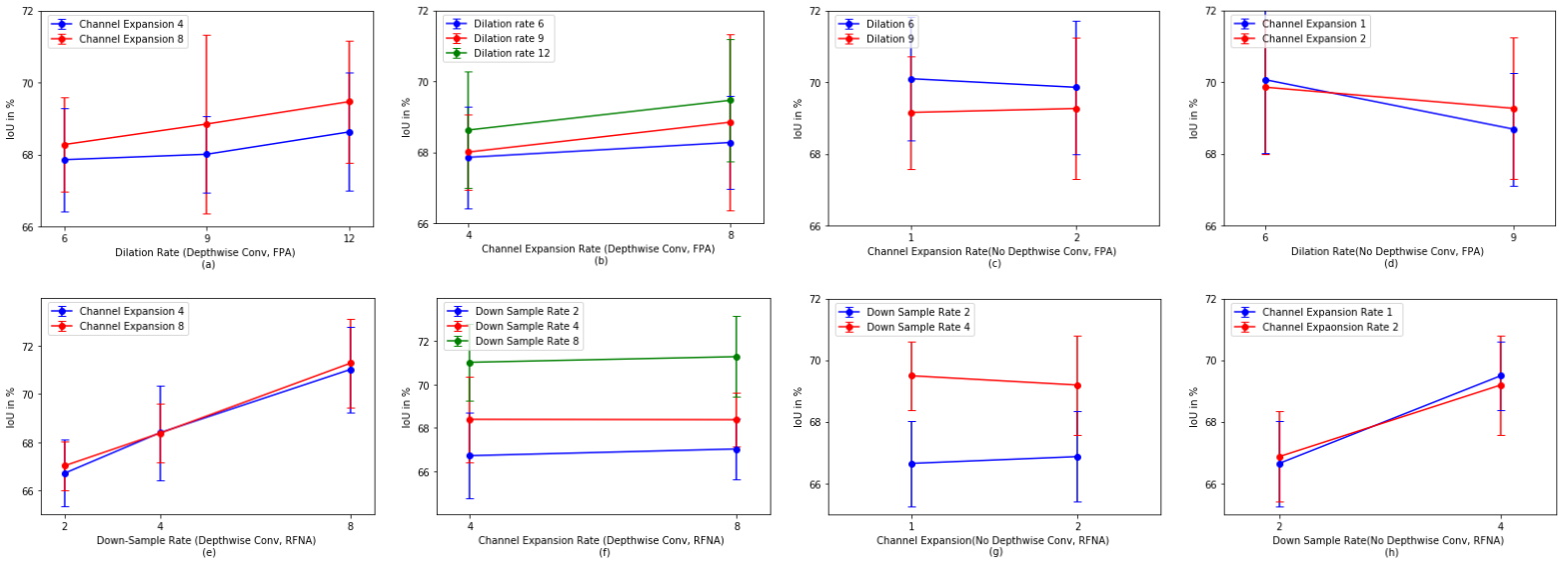}
  \vspace{-20 pt}
  \caption{Ablation studies of different configurations of FPA/RFNA on BRATS2018.}
\label{fig:hyperparameters}
\vspace{-15 pt}
\end{figure}

We found that in RFNA, a higher downsampling ratio consistently lead to better segmentation performance ((e) and (h) in \textbf{Figure}\ref{fig:hyperparameters}). However, the dilation rate in FPA only resulted in better performance when used with depth-wise convolutional layers ((a) and (d) in \textbf{Figure}\ref{fig:hyperparameters}). No clear relationships between channel expansion ratio and performance in either FPA or RFNA were observed, which might suggest that network architectures are more influential on performance than the parameters numbers.

\begin{table*}[!ht]
% \vspace{-10 pt}
\centering
\caption{Ablation studies of removing branches in FPA/RFNA on BRATS2018.}
\label{table:result_ablation}%
\centering
\begin{tabular}{l | c c c c c}
  \hline
  \bfseries Networks & \bfseries IoU ($\%$) & \bfseries HD & \bfseries FP ($\%$) & \bfseries FN ($\%$) & \bfseries Param (M)\\
  \hline
  Unet & 66.67 $\pm$ 1.15 & \textbf{14.77} $\pm$ \textbf{1.61} & $<$1e-2 & 57.83 $\pm$ 2.25 & 3.13\\
  WUnet & \textbf{66.89} $\pm$ \textbf{1.07} & 15.40 $\pm$ 0.87 & $<$1e-2 & \textbf{57.66} $\pm$ \textbf{3.20} & \textbf{12.51}\\
  D6Unet & 66.42 $\pm$ 1.14 & 15.174 $\pm$ 1.92 & $<$1e-2 & 58.52 $\pm$ 2.68 & 3.13\\
  D9Unet & 66.63 $\pm$ 1.41 & 15.61 $\pm$ 1.08 & $<$1e-2 & 58.07 $\pm$ 3.79 & 3.13\\
  \hline
  \end{tabular}
\vspace{-10 pt}
\end{table*}

\section{Conclusion}
\vspace{-10 pt}
\label{sec:conclusion}
In this paper, we present \textit{Pay Attention To Mistakes} to tackle the high FNs detection rate in medical image segmentation. Our method effectively reduces the FN rate of backbone and achieves superior performance compared to existing state-of-the-art attention mechanisms in medical image segmentation across three different data sets. Although both the FPA and RFNA implementations are shown to be effective on tasks suffering from over-detection of FNs, the use of RFNA on tasks suffering from over-detection of FPs will require exploration in future work. By flexibly apply FPA and RFNA we could potentially cover most of the situations where over-detection of FNs and over-detection of FPs happen in both medical and computer vision domains. 

\section{Acknowledgement}
Mou-Cheng is supported by GSK funding (BIDS3000034123) via UCL EPSRC CDT in i4health and UCL Engineering Dean's Prize. 

Neil is a UKRI Future Leaders Fellow (MR/S03546X/1) supported by the National Institute for Health Research University College London Hospitals Biomedical Research Centre. 

We thank NVIDIA for hardware donation. We also thank Fred Wilson from GSK; Eyjolfur Gdmundsson and Yi-Peng Hu from UCL CMIC for their feedback on the draft.
% Previous applications of existing attention mechanisms to medical image segmentation have focussed on direct implementation of tools from computer vision. This has resulted in limited improvements to reducing FNs, because the previous methodology does not explicitly take into account the commen class imbalance inherent to medical image segmentation tasks. Here we propose a novel attention mechanism that instead focusses attention on FPs. We show in experiments on the BRATS 2018 training dataset that our approach considerably outperforms nine state-of-the-art baseline methods in terms of Hausdorff Distance and intersection over the union. In future work, we plan to include two extensions. The first extension will evaluate the methods on an another dataset which is dominated by foreground pixels, whilst the second extension will propose a multi-class version of the current methods.
%------------------------------------------------------------------------- 
% \subsection{References}

% List and number all bibliographical references in 9-point Times,
% single-spaced, at the end of your paper. When referenced in the text,
% enclose the citation number in square brackets, for
% example~\cite{Authors06}.  Where appropriate, include the name(s) of
% editors of referenced books.

%------------------------------------------------------------------------
% \subsection{Color}

% Color is valuable, and will be visible to readers of the electronic copy.
% However ensure that, when printed on a monochrome printer, no important
% information is lost by the conversion to grayscale.

\bibliography{bmvc_final}
\end{document}